\documentclass[11pt,a4paper]{article}
\usepackage{times,latexsym}
\usepackage{url}
\usepackage[T1]{fontenc}

\usepackage[acceptedWithA]{tacl2018v2}

\usepackage{xspace,mfirstuc,tabulary}

\newif\iftaclinstructions{}
\taclinstructionsfalse{} 
\iftaclinstructions{}
\renewcommand{\confidential}{}
\renewcommand{\anonsubtext}{(No author info supplied here, for consistency with
TACL-submission anonymization requirements)}
\newcommand{\instr}
\fi

\iftaclpubformat

\else

\fi

\usepackage{amsmath}
\usepackage{amsfonts}
\usepackage{booktabs}
\usepackage{bm}
\usepackage{graphicx}
\usepackage{multirow}
\usepackage{subcaption}

\title{Adding Recurrence to Pretrained Transformers for Improved Efficiency and Context Size}

\author{Davis Yoshida\\Toyota Technological\\Institute at Chicago\\{\sf dyoshida@ttic.edu}
    \And
    Allyson Ettinger\\Department of Linguistics\\University of Chicago\\{\sf aettinger@uchicago.edu}
    \And
    Kevin Gimpel\\Toyota Technological\\Institute at Chicago\\{\sf kgimpel@ttic.edu}
}

\date{}

\begin{document}
\maketitle

\begin{abstract}
Fine-tuning a pretrained transformer for a downstream task has become a standard method in NLP in the last few years. While the results from these models are impressive, applying them can be extremely computationally expensive, as is pretraining new models with the latest architectures. We present a novel method for applying pretrained transformer language models which lowers their memory requirement both at training and inference time. An additional benefit is that our method removes the fixed context size constraint that most transformer models have, allowing for more flexible use. When applied to the GPT-2 language model, we find that our method attains better perplexity than an unmodified GPT-2 model on the PG-19 and WikiText-103 corpora, for a given amount of computation or memory.
\end{abstract}

\section{Introduction}

Recent progress in NLP has been dominated by large pretrained transformer neural networks~\cite{vaswani2017attention}, such as BERT~\cite{devlin-etal-2019-bert}, and GPT-2~\cite{gpt2}.
However, these models have a memory footprint that is quadratic in input sequence length.
Although architectural innovations such as those of \citet{reformer} and \citet{pg19} mitigate this and the issue of a predetermined maximum context size, large pretrained models applying these techniques are not available at this time.
Even if large pretrained models of this kind are released in the future, they will likely not cover the wide range of domains that BERT-family models have been published for. For example, there have been BERT-based models trained for other languages such as French~\cite{flaubert,camembert}, Italian~\cite{alberto}, and many other languages (see~\citet{bertlangoverview} for an overview) as well as specific domains such as scientific papers~\cite{scibert}, biomedical papers~\cite{biobert}, and health records~\cite{medbert}.
Individuals working with these models may not have the resources to train new models from scratch using the latest tricks, as the computation requirements for pretraining are extremely high. As such, identifying ways that already existing models can be improved could be widely impactful.

Another drawback of this family of models is that they have an a priori fixed maximum context size (typically 512 or 1024 tokens for the currently available pretrained models). A typical application of pretrained language models is producing contextual embeddings for a document.
If the document is simply chunked into disjoint segments of 512 tokens, tokens at the boundary of a window will have less contextual information than tokens in the center of a window.
This can be mitigated by striding the evaluation of the model, and only keeping the embedding for a token which has the largest context---but this adds quite a bit of wasted computation.

In this paper, we propose a method for augmenting and fine-tuning pretrained transformer language models to use context without directly attending to it.
Our method simultaneously allows for increasing the context size a transformer processes, while allowing a controllable trade-off between computation and perplexity.
We accomplish this by adding a small recurrence module that computes a fixed size representation from the transformer hidden states in a window of text.
Then, the representation for that window is used during processing of the next window.
Shrinking the window size is then a way to reduce the memory footprint of the model, with less loss of performance than would occur with a standard transformer.
Our experiments add recurrence GPT-2 language models, and fine-tune them on the PG-19~\cite{pg19} and WikiText-103 corpora~\cite{wikitext103}, and require only the same amount of memory used for standard fine-tuning of a pretrained language model.
We demonstrate improvements in perplexity compared to a baseline model using the same amount of computation.
Qualitative analysis shows that our recurrent module propagates certain information from previous windows of text, which can facilitate handling of long-distance dependencies with fixed-size input windows.

\section{Related Work}
Recently, many methods have been proposed which lower the memory footprint or computation time of transformer language models, or allow them to be used on larger contexts.
The Transformer-XL~\cite{dai-etal-2019-transformer} allows a position within an attention window to attend to tokens from the previous windows by introducing relative position embeddings.
While that mechanism like ours, allows information to flow between windows of text, existing BERT and GPT-2 models do not use relative position embeddings, so training from scratch would be necessary to take advantage of this architecture.

Other methods modify the attention function to reduce the quadratic memory footprint down to a manageable amount.
\citet{child2019generating} modify the transformer architecture to replace the standard attention with a sparse one.
\citet{qiu2019blockwise} enforce a block-sparse structure on the attention matrix.
\citet{reformer} also introduce sparsity, but instead do so by using locality sensitive hashing to select positions over which a full attention is computed, reducing the memory cost from quadratic to $O(T\log{T})$ for an input of size $T$.
\citet{pg19} introduce a memory compression technique that allows much longer contexts to be attended to in memory.
\citet{longformer} replace the standard attention with a combination of dilated sliding windows, and global attention from selected tokens that.
\citet{sukhbaatar-etal-2019-adaptive} learn a masking function such that not all tokens attend to every previous position.
\citet{Tay2020SynthesizerRS} learn synthetic attention weights, removing the need for token-token interactions.
\citet{Wu2019PayLA} replace the full self-attention with a dynamic convolution depending only on the current timestep, yielding a linear dependence on length instead of a quadratic dependence.

While the above methods all allow for a reduction in required computational resources, they also all require one to train a model from scratch. Our method's goal is to allow more efficient and powerful use of the wide array of existing pre-trained models that cover many domains.

\citet{Cao2020DeFormerDP} propose the DeFormer, which also modifies the execution of a pretrained transformer. However, unlike our method, they decompose a single window into multiple windows by removing the attention interactions between these windows. This is largely orthogonal to our method, as one could both decompose windows of text, and additionally use our method to allow information to be passed between neighboring windows. Similarly, distilled versions of pretrained models such as DistilBERT~\cite{distilbert} provide more computational efficiency, but could be combined with our method to apply them to longer contexts, or reduce the quadratic cost of self-attention.

\section{Method}\label{method}
\begin{figure}
    \centering
    \includegraphics[width=\columnwidth]{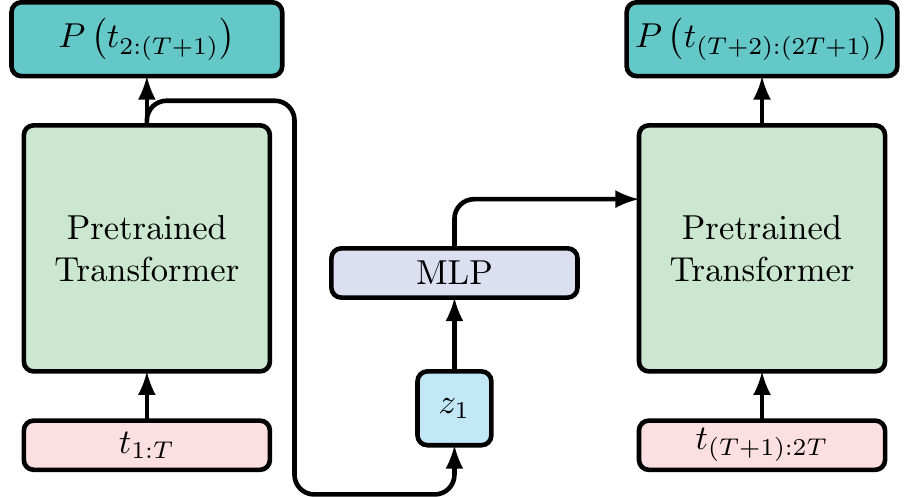}
    \caption{\small Augmenting a pretrained transformer with a small recurrence module, allowing reduction of attention computation as well as simpler processing of longer contexts.}\label{fig:model}
\end{figure}

The main idea of our method is to take a transformer that was pretrained in a fixed context size setting and add recurrence at the level of $T$-token windows of text.
For example, instead of executing the model on one 1000 token window of text, we could instead execute our model with 10 windows of 100 tokens.
The first window is processed by the transformer model as normal, but for subsequent windows we add a supplementary embedding, which is generated using the hidden states from the preceding window (see Figure~\ref{fig:model}).
The recurrence module is extremely small compared to the size of transformer language model, so the additional computation required is negligible.

\subsection{Adding recurrence to pretrained transformers}
Starting by defining terms, we will consider a pretrained transformer with $L$ layers, a hidden state size of $k$, and a maximum context size of $T$ tokens. Let $\bm{h}_i^{(\ell)} \in \mathbb{R}^{k}$ be the output of the $\ell$-th layer of the pretrained model, at position $t$. To produce a fixed-size representation of tokens $t_1, t_2, \dots, t_T$, the embeddings produced by the pretrained transformer are mean-pooled as follows:
\begin{equation}\label{eq:pool}
    \bm{z}_1 = \sum\limits_{i=1}^T \sum\limits_{\ell=1}^L w_\ell \bm{h}_i^{(\ell)}
\end{equation}
where $w_\ell$ are weights softmax-normalized from learned parameters $\alpha_\ell$:
\begin{equation*}
    w_\ell = \frac{\mathrm{e}^{\alpha_\ell}}{\sum\limits_{j=1}^L \mathrm{e}^{\alpha_j}}
\end{equation*}

\begin{figure}\label{fig:token_insertion}
    \centering
    \includegraphics[width=\columnwidth]{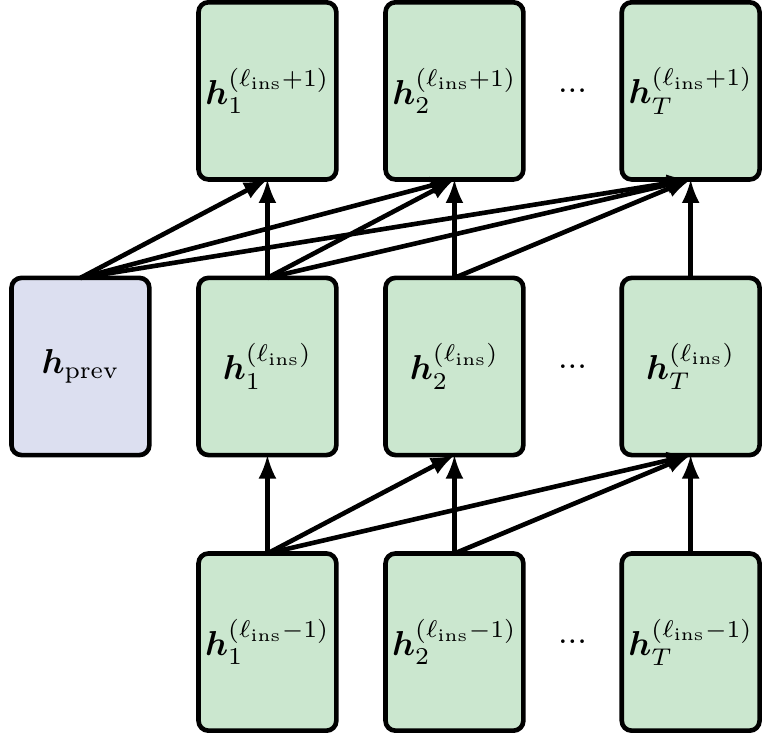}
    \caption{\small $\bm{h}_{\text{prev}}$ is added as an additional key and value to one self-attention layer. Arrows show which positions can pass information to which other positions.}\label{fig:token_insert}
\end{figure}
The fixed-size representation, $\bm{z}_1$, is passed through a feedforward network to produce an embedding $\bm{h}_{\text{prev},1}$ which represents the tokens processed so far, $t_{1:T}$. Next, instead of evaluating the pretrained transformer without modification on positions $T+1$ through $2T$, $\bm{h}_{\text{prev},1}$ is inserted at a single layer (denoted $\ell_{\text{ins}}$) of the pretrained model, as an additional embedding that may be used in the computation of attention, as shown in Figure~\ref{fig:token_insert}. To keep the number of embeddings per layer fixed, this embedding is only used as a key and a value, but not a query, in the self-attention layer.
That is, for a window size of 300 tokens, there are 301 inputs to layer $\ell_{\text{ins}}$, but still only 300 outputs.
The embeddings for positions $T+1$ to $2T$ are then pooled in the same way as Equation~\ref{eq:pool} to produce $\bm{z_2}$ and passed through the feedforward network, outputting $\bm{h}_{\text{prev},2}$. $\bm{h}_{\text{prev},2}$ is used to modify the execution of the pretrained language model on tokens $2T +1$ through $3T$, and so on. Because the model is now being applied recurrently, it is trained end-to-end with backpropagation through time.

One could consider more complex recurrence modules, different methods for pooling the previous window's embeddings, or for inserting $\bm{h}_{\text{prev}}$ into the computation for the next window. We experimented with modifications such as max pooling instead of mean pooling, inserting multiple embeddings into the next window, inserting an embedding at all layers of the transformer for the next window, and using fixed key attention as the pooling function. However the performance after each of these changes was not significantly better than the model presented above, so we do not include those modifications here.

\subsection{Gradient checkpointing in networks with bottlenecks}
While our method can reduce the quadratic cost of attention by splitting the input into windows, we can also easily apply it to much longer contexts by use of gradient checkpointing~\cite{chen_checkpointing}.

Gradient checkpointing is a method for lowering the peak memory requirement of training large neural networks. This is accomplished by storing only a subset of activations during the forward pass, and recomputing forward from those cached states during the backwards pass. For example, in a 100 layer feedforward network with uniformly wide layers, one could store the output of only every 10th layer. Then, during the backward pass, in order to compute the gradients for the 95th layer, one would re-compute layers 91 through 99 using the stored 90th layer activations. The overall memory cost is reduced to $\sqrt{L}$ at the cost of a single additional forward pass.

In a network with variable width, the memory reduction can be even larger. When gradient checkpointing is applied to transformers, the outputs of each layer are usually stored ($k\times L \times T$ values), so that at most one set of self-attention activations is stored in memory at once. In the case of our recurrent models, we have an even narrower bottleneck: the $\bm{z}_i$'s and $\bm{h}_{\text{prev},i}$'s. Storing only these values means that the maximum number of activations present in memory while training on sequences $N$ tokens in length is $M + 2k\lceil{\frac{N}{T}}\rceil$, where $M$ is the number of activations stored when training the transformer on an individual window of length $T$. Because $k$ is extremely small compared to $M$, our model can be applied to very long contexts on any GPU on which the pretrained model could be fine-tuned.

\section{Revisiting the evaluation of transformer language models}\label{transformer_eval}
Before describing the empirical evaluation of our method, we discuss how transformer language models are evaluated in related work. The standard way of measuring perplexity uses extra computation in order to make as much context available for each token prediction. This yields low perplexities, but does not reflect how practitioners use transformer language models in applications. In this section, we describe the situation in detail and propose practical solutions that achieve relatively low perplexities while being closer to how transformers are used in practice.

\subsection{Potential misalignment between LM evaluation and application}
Transformers are often described as having quadratic time complexity in comparison to RNNs which have linear time complexity. However, this can be somewhat misleading when it comes to evaluation of perplexity on a test set. Given a test set of length $N$, an RNN requires $O(N)$ time to evaluate---but reaching the best perplexity for a transformer requires $O(NT^2)$, where $T$ is its maximum context size. (The preceding time complexities exclude hidden state size, number of layers, and batch size.)
    This much higher time complexity is due to the fact that a transformer may be run with its full context size once for each token in the test set, so that the maximum context is available for each prediction.
    Re-execution of the whole model for each token is required for models with absolute position embeddings, since hidden state reuse is only possible up to the maximum context size of the network.
    Note that it is possible to achieve smaller wall-clock time by splitting evaluation of a test set over multiple GPUs, but this is not applicable to the generation setting where outputs depend on prior ones.

\begin{figure*}
    \centering
    \begin{subfigure}{0.3\textwidth}
        \centering
        \includegraphics[width=0.8\linewidth]{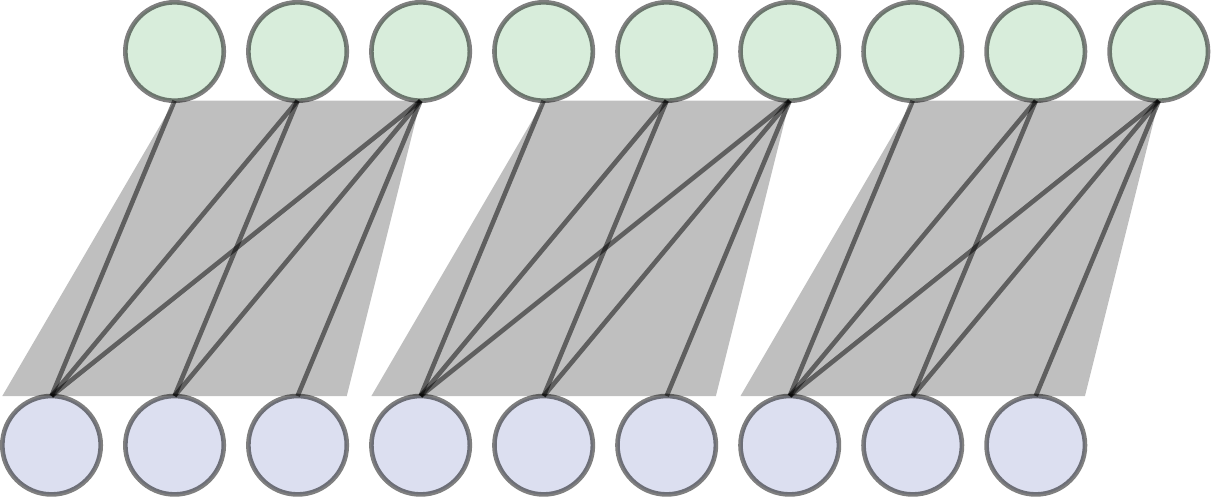}
        \caption{\small Disjoint execution. Predictions have context ranging between 1 and 3 tokens.}\label{fig:disjoint}
    \end{subfigure}
    \hfill
    \begin{subfigure}{0.3\textwidth}
        \centering
        \includegraphics[width=0.8\linewidth]{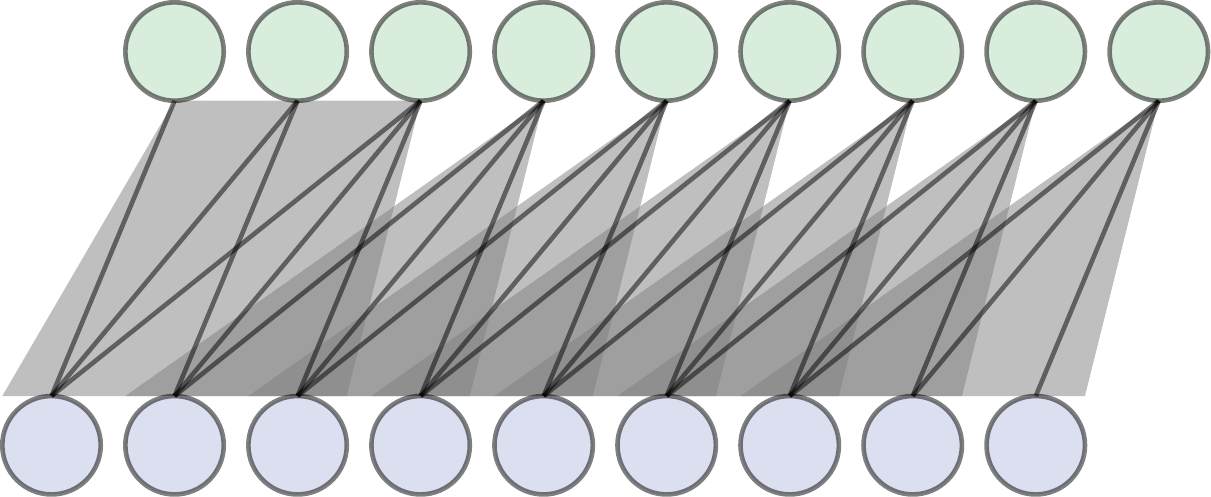}
        \caption{\small Maximum overlap. All predictions except the first two have maximal context. }\label{fig:max_overlap}
    \end{subfigure}
    \hfill
    \begin{subfigure}{0.3\textwidth}
        \centering
        \includegraphics[width=0.8\linewidth]{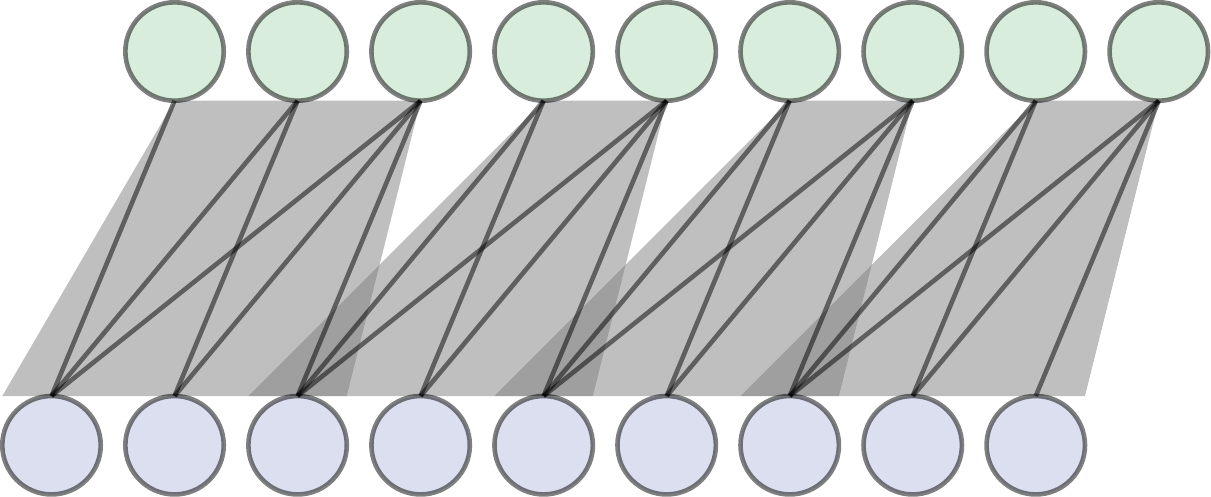}
        \caption{\small Intermediate degree of overlap. Except the first prediction, all predictions attend to at least 2 tokens of context.}\label{fig:partial_overlap}
    \end{subfigure}
    \caption{\small Varying degree of overlap while evaluating a transformer with a window size of 3. The green (top) circles are outputs, and the blue (bottom) circles are inputs.\label{fig:overlap}}
\end{figure*}

    To illustrate why re-computation is necessary, consider executing GPT-2 (which has 1024 position embeddings) on a test set.
    Each of the first 1024 tokens of a test set will have been passed into the network using a distinct position embedding.
    Having exhausted the position embeddings, one option is to start again with the 1025th token being treated as position 1---we will refer to this as \emph{disjoint execution}, illustrated in Figure~\ref{fig:disjoint}. The issue with disjoint execution is that it requires predicting the tokens at the beginning of a window from a very small amount of context.

    The alternative, which is used for standard test set evaluation, is \emph{overlapped execution}, as shown in Figure~\ref{fig:max_overlap}.
    The position embeddings are advanced by one position for each prediction, meaning that $T-1$ tokens are repeated between consecutive evaluations of the transformer, requiring much more computation.
    The benefit of this method is that it allows a model with $T$ position embeddings to have $T$ tokens of context for each prediction, as opposed to a variable amount between 1 and $T$.

    Stepping a transformer decoder forward one token at a time measures the best that such a model could perform, but it reflects a generative story that does not align with how the models may be used in practice.
    A perplexity that only measures the ability of GPT-2 to generate the 1024th token given a context of 1023 tokens is not necessarily indicative of the model's performance when generating from a smaller context.
    For example, the popular website Talk To Transformer\footnote{https://talktotransformer.com/} generates samples from GPT-2, but only provides 150 tokens of output.
    The evaluation of GPT-2 by stepping forward one token at a time provides little information about the quality of such generations.

    An example where the discrepancy is length instead of brevity is the GPT backed text adventure game AI Dungeon.\footnote{https://aidungeon.io/. Note that AIDungeon now uses the OpenAI GPT-3 API, but a similar project without OpenAI API access would still have to use GPT-2.} In this setting, the number of tokens can easily reach and exceed the full context size GPT-2 was trained on.
    Using overlapped execution as described above, generating each token would be 1024 times slower than with disjoint execution, so perplexity calculated by overlapped execution does not match this use case either.

While lower perplexity seems to correspond to better generation with shorter contexts in practice (perhaps due to parameter sharing between all sequence positions), there is no reason that this need be the case in principle. To demonstrate an extreme case of the concern being discussed, let $F$ be a transformer model with vocabulary $V$, which uses the previous 1023 tokens as context, and consider the following generative story for generating token $t_i$:
\begin{equation*}
    t_i \sim \begin{cases}
        \text{Uniform}(V) & \text{if $i \le 1023$}\\
        F(t_{(i-1023):(i-1)}) & \text{otherwise}
    \end{cases}
\end{equation*}
Clearly the above generative model would not be of any practical use for generation or otherwise. However, because perplexity is calculated per token, increasing the size of the test set will lead to a measured perplexity that approaches that of a standard evaluation of the model $F$.
This example is not representative of the models that are trained in practice, as even generations much shorter than the maximum context size from a GPT-2 model are quite impressive.
However, it does demonstrate that the criteria that we use to compare models, or to select the best model during early stopping, place very high weight on the ability of the model to produce text given a full context, and a potentially vanishingly small amount on its ability to generate text using shorter contexts.

\subsection{Varying overlap for evaluation}\label{overlap}
As we are interested in increasing computational efficiency at evaluation time for pretrained models, we investigate their performance using overlapped execution, but with a reduced degree of overlap between windows.
Varying the overlap lets us investigate the connection between degree of overlap and perplexity.
The overlap used in evaluation will be defined to be the number of tokens from each input window that are repeated in the next window (see Figure~\ref{fig:overlap}).
For example, consider a window size $T = 10$, and an overlap of 3.
The windows that the transformer will be executed are then $t_{1:10}$, $t_{8:17}$, $t_{15:24}$, \textellipsis, $t_{1 + 7n:10 + 7n}$where $n$ indexes the window.
These input windows are used to predict the spans of tokens $t_{2:11}$, $t_{12:18}$, $t_{19:25}$, \textellipsis, $t_{5 + 7n:11+ 7n}$.
Figure~\ref{fig:partial_overlap} illustrates an intermediate overlap setting with $T=3$, and an overlap of 1.
The perplexity minimizing evaluation setting for a transformer is then the extreme with an overlap $T-1$, and an overlap of 0 corresponds to disjoint execution.

While a standard transformer can be evaluated with any degree of overlap, our augmentation method produces the embedding $\bm{h}_{\text{prev}}$, which is used during training to help predict the first token of a window of text. If we change the overlap at test time, the alignment of the text represented by $\bm{h}_{\text{prev}}$ and the current window of text will be different than the model was trained for, and so performance will degrade. To address this, we use the same overlap that will be used at test time during training for the recurrent models.

\section{Experiments}
In this section we provide experiments comparing our proposed technique to the default usage of transformer language models. We describe experiments on the WikiText-103 corpus and a subset of the PG-19 corpus, using the GPT-2-small language model as the pretrained transformer component of our models.

WikiText-103 is a standard corpus for word level language modeling composed of approximately 29,000 documents from English Wikipedia, totaling about 103 million words. We use the WikiText-103 ``raw'' corpus, which does not have rare words replaced by ``UNK'' tokens. While we train using the BPE tokenization used by GPT-2, we normalize the final loss by the number of words rather than the number of BPE tokens for clarity.

Although WikiText-103 does test long term dependencies, many of the documents are still shorter than the context size of the models we test. Therefore, we also use PG-19, which consists of books from the Project Gutenberg corpus. The average length of a WikiText-103 document is 3.6K words, while PG-19 documents (i.e.\ books) average 69K words, which far exceeds the context size of the models we test. However, the full PG-19 dataset is over 20 times larger than WikiText-103, so we use only a subset of it for training due to computational constraints. Specifically, we use only the first (alphabetically by filename) 1250 books of the PG-19 corpus, and use only the first 15000 tokens of each of the books in the validation set for early stopping. We make no modifications to the test set.

In all our experiments we use the HuggingFace implementation of the pretrained GPT-2 small model (12 layers, 768-dimensional hidden state). For both the recurrent and baseline models, the GPT-2 model was fine-tuned, not left frozen. We selected learning rates for both our models and the baseline separately, by evaluating on WikiText-103 for the same set of candidate learning rates. We used the same learning rates for the PG-19 experiments without further hyperparameter search. We fine-tune all models for 2 epochs, measuring the validation loss every 2 million tokens. All models were trained with Adam~\cite{adam}, warming the learning rate up linearly from 0 to its final value over 100 steps. The feedforward network used to produce $\bm{h}_{\text{prev}, i}$ from window $i-1$ consisted of 3 hidden layers with dimension 200. We fixed $\ell_{\text{ins}}$ to be 2.

Recall from Section~\ref{transformer_eval} that we are interested in evaluating the models in a setting similar to how they would be used in practice. To that end, we report separate perplexities for different degrees of overlap between adjacent windows of text, as described in Section~\ref{overlap}. For our models, we train with the same overlap that we test with, as unlike the baseline models, they cannot be trained with no overlap between adjacent windows and then tested with an overlap. This is because the embedding of the previous window of text is expected to represent all tokens up until the first token of the current window, but with an overlap of 30 for example, that embedding would be representing all tokens up until the 30th token of the current window.

\subsection{Results}
We first show that with the same amount of fine-tuning, our method achieves lower perplexity than a baseline GPT-2 model when evaluated using the same window size and degree of overlap between adjacent windows of text.

It is important to emphasize that the perplexities we report are based on pretrained models, and so should not be compared to models trained from scratch on these datasets. The GPT-2 models were trained on text from a web crawl from which all Wikipedia documents are removed, but this still leaves open the possibility of quotes from Wikipedia having been encountered, or text from PG-19.

\begin{table*}
\small
\centering
\begin{tabular}{p{3cm}cp{2cm}p{3cm}c}
    Model & Overlap & Validation Perplexity & Test Perplexity & FLOPs/token\\
    \midrule
    \multirow{5}{3cm}{GPT-2 (small), 300 token window} & 0 & 29.00 & 30.47 & $1.75\times 10^{8}$\\
    & 5 & 27.99 & 29.36 & $1.78\times 10^{8}$\\
    & 10 & 27.58 & 28.88 & $1.81\times 10^{8}$\\
    & 30 & 26.72 & 27.96 & $1.94\times 10^{8}$\\
    & 50 & 26.17 & 27.31 & $2.10\times 10^{8}$\\
    \midrule
    \multirow{5}{3cm}{Recurrent, 20 windows of 300 tokens (Ours)} & 0 & 27.70 & 29.01 & $1.75\times 10^{8}$\\
    & 5 & 26.88 & 28.12 & $1.78\times 10^{8}$\\
    & 10 & 26.51 & 27.77 & $1.81\times 10^{8}$\\
    & 30 & 25.90 & 27.12 & $1.94\times 10^{8}$\\
    & 50 & 25.53 & 26.73 & $2.10\times 10^{8}$\\
\end{tabular}
\caption{Results on WikiText-103}\label{tab:wikitext}
\normalsize
\end{table*}

Table~\ref{tab:wikitext} shows the perplexity of our models and the non-recurrent GPT-2 models on the WikiText-103 dataset.
The models compared here all use windows of 300 tokens, with varying degrees of overlap.
The baseline models can only access information from the previous window of text through the overlapping tokens, while the recurrent models have a fixed size representation of the longer context.
Our addition of recurrence increases the performance of the GPT-2 models in this setting, but by a relatively small amount.
Increasing the overlap between each window of text decreases the perplexities of the baseline model as expected, but also decreases the perplexity of the recurrent models.\footnote{We did not attempt to train recurrent models with extremely high overlaps, as that would greatly increase the required training time}
This indicates that there is room to increase the capacity of the recurrence mechanism, as if it passed all relevant information about the previous window forward, the overlapping tokens would be redundant.
On the other hand, some useful information beyond what is contained in the local context is being propagated, as otherwise the baseline model should catch up in perplexity at higher overlaps.
To investigate this further, we also experiment with the PG-19 dataset.

The results for the PG-19 experiments are shown in Table~\ref{tab:pg19}.
While we find only small increases in performances on the WikiText-103 dataset, we see larger improvements on PG-19, confirming our prediction that the gains would be larger on a dataset that has a larger context available for each prediction on average.
We find that adding our recurrence module leads to a model that gives as good a perplexity with no overlap between adjacent windows as an unmodified model does when evaluated with an overlap of 30 out of 300 tokens in each window.
Training the recurrent model with a 5 token overlap gives perplexity lower than the baseline perplexity with an overlap of 50 or even 75. In terms of FLOPs, adding our recurrence module and overlapping adjacent windows of tokens by 50 is less than half as costly as using a non-recurrent model with an overlap of 200.

\begin{table*}[t]
\small
\centering
    \begin{tabular}{p{3cm}cp{2cm}p{3cm}c}
    Model & Overlap & Validation Perplexity & Test Perplexity & FLOPs/token\\
    \midrule
        \multirow{9}{3cm}{GPT-2 (small), 300 token window} & 0 & 172.25 & 147.71 & $1.75\times\mathrm{10}^8$\\
        & 5 & 165.93 & 142.30 & $1.78\times 10^8$\\
        & 10 & 162.66 & 139.49 & $1.81\times 10^8$\\
        & 30 & 156.21 & 134.30 & $1.94\times 10^8$\\
        & 50 & 152.64 & 131.25 & $2.10\times 10^8$\\
        & 75 & 149.54 & 128.46 & $2.33\times 10^8$\\
        & 100 & 147.05 & 126.51 & $2.62\times 10^8$\\
        & 150 & 143.62 & 123.53 & $3.50\times 10^8$\\
        & 200 & 141.14 & 121.40 & $5.25\times 10^8$\\
    \midrule
        \multirow{5}{3cm}{Recurrent, 20 windows of 300 tokens (Ours)} & 0 & 155.27 & 133.02 & $1.75\times 10^8$\\
        & 5 & 150.00 & 128.78 & $1.78\times 10^8$\\
        & 10 & 147.53 & 127.05 & $1.81\times 10^8$\\
        & 30 & 142.35 & 122.22 & $1.94\times 10^8$\\
        & 50 & 140.10 & 119.93 & $2.10\times 10^8$\\
\end{tabular}
\caption{Results on PG-19}\label{tab:pg19}
\normalsize
\end{table*}

\subsection{Effect of window size}
\begin{figure}
\centering
    \includegraphics[width=\columnwidth]{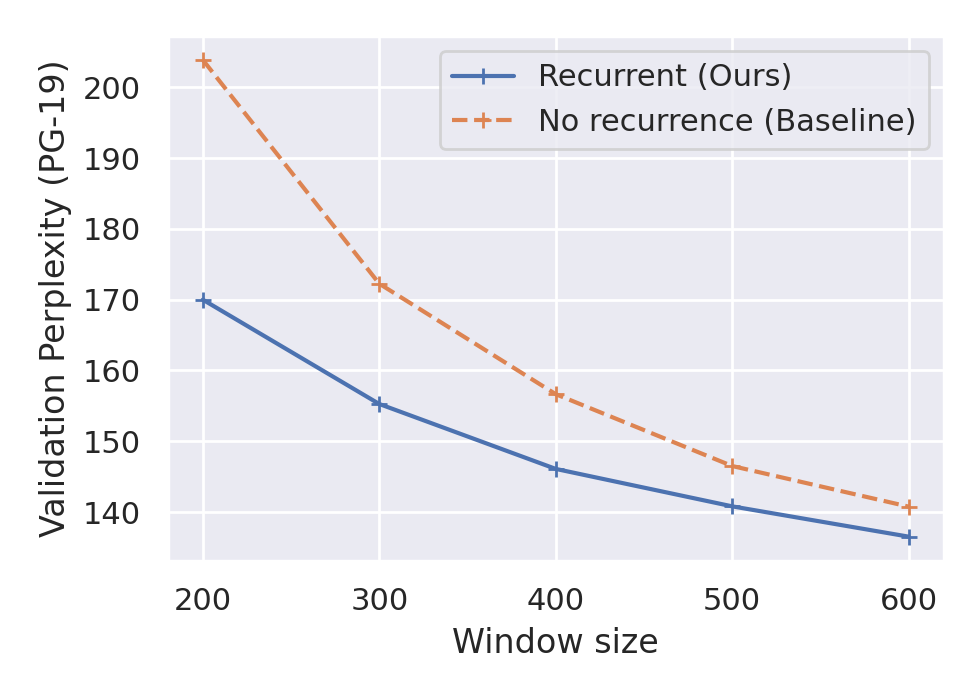}
\caption{\small Effect of window size on performance on PG-19 validation set}\label{fig:window_size}
\end{figure}

\begin{figure}
\centering
    \includegraphics[width=\columnwidth]{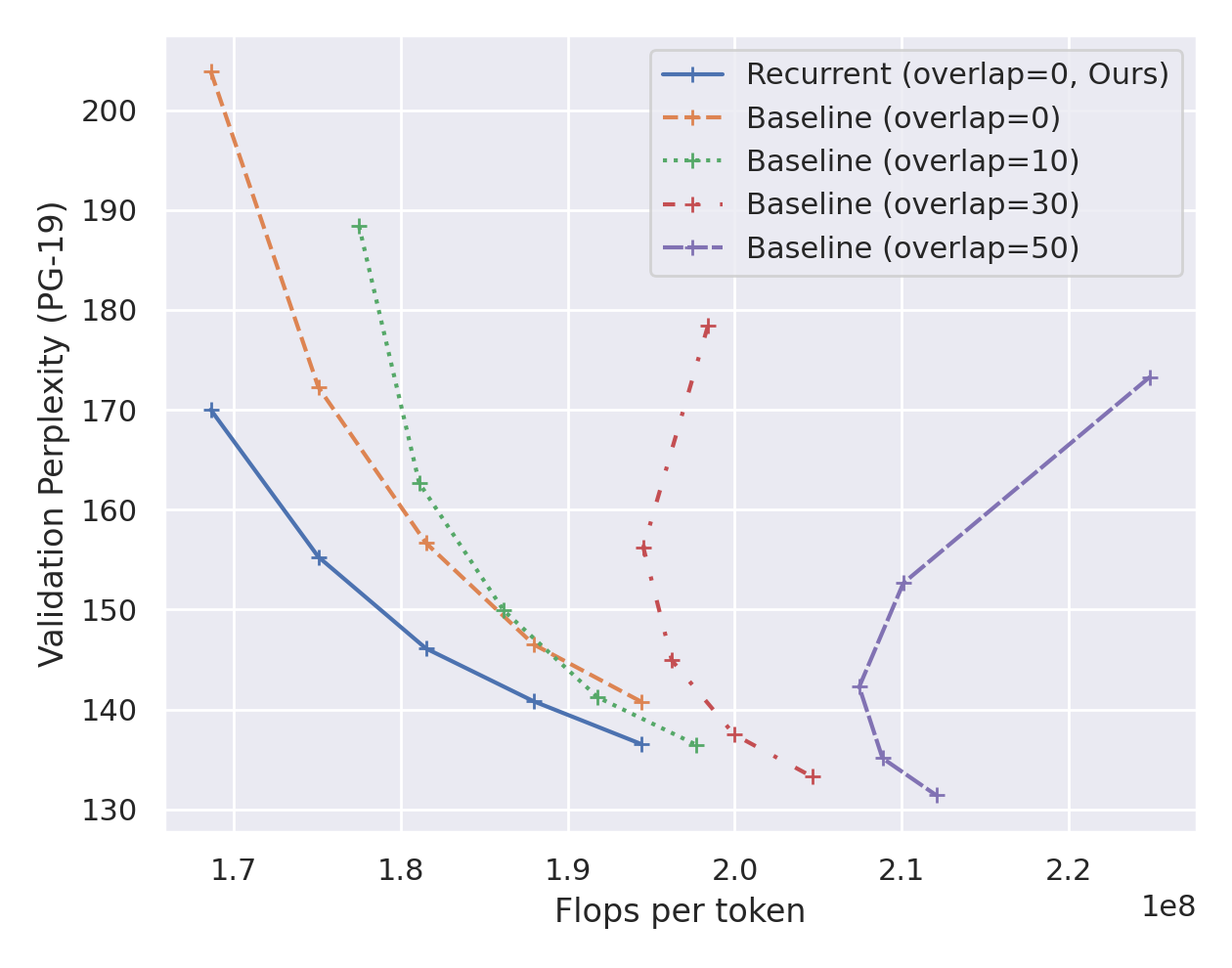}
\caption{\small Relationship between FLOPs and perplexity for recurrent and non-recurrent models. Each curve ranges over window sizes from 200 to 600.}\label{fig:flops}
\end{figure}

As one of our motivations is to retain performance while decreasing compute requirements, we experiment with varying the window size used by our model and an unmodified GPT-2 model. At smaller window sizes the recurrent model has access to much more information than an unmodified GPT-2 model, which can only attend to the current window of tokens. Because of this, we expect our augmentation to cause the performance to fall off less rapidly with decreasing window size. The results, shown in Figure~\ref{fig:window_size}, confirm this prediction, as the performance gap widens with smaller window sizes. Figure~\ref{fig:flops} contains the same points (and additional baseline curves for various overlaps), but in terms of FLOPs rather than window size. In this comparison, all of the results of the recurrent models lie on the Pareto frontier, meaning that to improve perplexity or computational cost, one must worsen the other. The non-monotonicity of the overlap 30 and 50 curves is due to the fact that at smaller window sizes, an overlap represents a higher fraction of the computation being used for positions that predictions were already produced for. Also note that while the baseline with overlap 50 curve has the lowest absolute perplexity in Figure~\ref{fig:flops}, the recurrent models trained with overlaps shown in Table~\ref{tab:pg19} still perform better.

\subsection{What information is being propagated between windows?}
\begin{table*}[t]
    \small
    \setlength{\tabcolsep}{4pt}
    \begin{tabular}{p{0.48\textwidth}p{0.37\textwidth}p{0.09\textwidth}}
        Context & Continuation & Topic\\
    \midrule
        \textellipsis{} It drains a territory of 300,000 square miles, and, traced back to the rise of its principal source, is 2,000 miles long. At two points, Needles and Yuma on the California boundary, it is crossed by a railroad. Elsewhere its course lies far from Caucasian settlements and far from the routes of common travel, in the heart of a vast region fenced on the one hand by arid plains or deep forests and on the other by formidable mountains & .  The first of the two great rivers, the Mississippi, was the first to be navigated by the Americans, and the first to be navigated by the Indians. The other was the first to be navigated by the Americans, and the first to be navigated by the Indians. The first of the two great streams, the Mississippi, was navigable by the Indians, and the first by the English\textellipsis{} & Rivers \\
    \midrule
        \textellipsis{} ``Mr. Hammerby will have to wait for his money.'' ``Will he?'' ``I don't see what else he can do.  But, aside from that, three dollars and a half won't keep us a month.  I'll have to &  get the money, and he'll be glad to see you again.'' ``I'm glad to see you again, Tom. I'm glad to see you again. I'm glad to see you again. I'm glad to see you again. I'm glad to see you again & Money\\
    \midrule
        \textellipsis{} Long's his sister, Thankful, lived, we couldn't say nothing, of course. If she wanted to slave and work so's her brother could live in idleness and sloth, why, that was her business. There ain't any law against a body's making a fool of herself, more's the pity. But she's been dead a year, and he's done nothing since but live on those that'll trust him, and ask help from the town. He ain't sick--except sick of work. [\textellipsis{}] to-mor & rer and his wife, and the whole family.  He was a good man, and he was a good man.  He was a good man, and he was a good man.  He was a good man, and he was a good man.  He was a good man, and he was a good man.  He & Family, Description of a man\\
    \midrule
        \textellipsis{} and any form of thought or any institution tending to suppress education or destroy intelligence strikes at the very essence of the government, and constitutes a treason which no law can meet, and for which no punishment is adequate. Education, then, as universally diffused as the elements of God, is the life-blood of our body politic. The intelligence of the people is the one great fact of our civilization and our prosperity,--it & is the only way to preserve the peace of the world. The only way to preserve the peace of the world is to preserve the liberty of the people.  The first step to be taken is to abolish the power of the government. The first step to be taken is to abolish the power of the people. The first step to \textellipsis{} & Government\\
    \midrule
\end{tabular}
    \caption{\small Continuations of PG-19 validation set passages using recurrent model ($T=300$, disjoint) which display approximate continuation of topic. Contexts are shortened for space, 300 tokens of context were used for each.}\label{tab:samples}
    \normalsize
\end{table*}

\begin{table*}
    \small
    \setlength{\tabcolsep}{4pt}
    \begin{tabular}{p{0.57\linewidth}p{0.39\linewidth}}
    Context & Continuation\\
    \midrule
        \textellipsis{} Elaine, however, did not gratify their hopes; she referred to the ball with the detachment she would have shown in describing a drawing-room show of cottage industries.  It was not difficult to discern in her description of the affair the confession that she had been slightly bored.  From Courtenay, later in the day, the aunts received a much livelier impression of& the house, and the two women were standing in the doorway. ``I am afraid I have not been very well,'' she said, as she looked at the two men.  ``I have been very much worried about you, and I have been very much worried about you.  I have been very much worried about you, and I have been very much worried about you.  I have been very much pleased with you, and I have been very much\\
    \midrule
        \textellipsis{} ``Oh, he's not at all a bookworm,'' said Suzette quickly, ``though he's tremendously well-read.  He's quite the man of action.'' ``Does he hunt?'' asked Elaine. ``No, he doesn't get much time or opportunity for riding.'' ``What a pity,'' commented Elaine; ``I don't think I could marry a man who wasn & '' ``I don't know,'' said the Sergeant, ``but I have heard that you are a very good friend.'' ``I am not,'' said the girl, ``I am not.'' ``I am not,'' said the Sergeant, ``I am not.'' ``I am not,'' said the girl, ``I am not.\\
    \midrule
        \textellipsis{} Gregg was still the soft-stepping scoundrel who invited the toe of honesty, and Mornway, as he entered, was conscious of a sharp revulsion of feeling. But it was impossible to evade the interview, and he sat silent while the man stated his case.  Mrs. Mornway had represented the stenographer as being in desperate straits, and ready to accept any job that could be found, but though his appearance might have seemed to corroborate her account, he evidently took a less hopeless view of his case, and the Governor found with surprise that he had fixed his eye on a clerkship & that he had been in the habit of doing, and that he had been in the habit of doing.  ``I'm not going to be a fool,'' he said, ``and I'm not going to be a fool,\textellipsis{}\\
    \midrule
        \textellipsis{} Many transactions too, from a desire to avoid worrying me, were carried through without my knowledge, although formerly, as a matter of course, they would have been submitted to me.  Strangers, when they called, asked to see Johnson or Marsh.  I directed the messenger that they were to be shown into my room if I was disengaged.  This was a failure, for, when they came, I was obliged to ask for help, which was not given very generously.  Sometimes I sent for the papers, but it took a long time to read them, and my visitors became impatient.  During one of these interviews, I remember that I was sorely perplexed, but I had managed to & get the money, and I was glad to get it.  The next day I went to the Court of the Emperor, and found him standing in the courtyard of the Palace, with a large number of soldiers.  He was dressed in a very dirty uniform, and wore a very dirty hat.  He was dressed in a very dirty uniform, and wore a very dirty hat.  He was dressed in a very dirty hat, and wore a very dirty hat\\
    \midrule
        \textellipsis{} I know at which crossings to look out--I know what I'm going to see in the shop-windows. It saves a lot of wear and tear to know what's coming. For a good many years I never did know, from one minute to another, and now I like to think that everything's cut-and-dried, and nothing unexpected can jump out at me like a tramp from a ditch.'' He paused calmly to knock the ashes from his cigar, and Garnett said with a smile: ``Doesn't such a plan of life cut off nearly all the possibilities?'' The old gentleman made a contemptuous motion. ``Possibilities of what? Of being multifariously miserable? There are lots of ways of being & ''I'm afraid I'm not going to be able to do that,'' he said. ``I'm going to have to go to the station. I'm going to have to go to the station. I want to see the station. I want to see the station. I want to see the station. I want to see the station. I want to see the station. I want to see the station. I want to see the station. I want to see\\
    \midrule
\end{tabular}
    \caption{\small Continuations of PG-19 validation set passages using recurrent model ($T=300$, disjoint) which display continuation of correct pronouns or references to characters from context. Contexts are left-truncated, 300 tokens of context were given for each continuation.}\label{tab:pronouns}
    \normalsize
\end{table*}

In this section we show sample continuations (using greedy argmax decoding) from contexts in the PG-19 validation set, which illustrate types of information that the recurrent module passes (or fails to pass) forward.
Table~\ref{tab:samples} shows that high level topical information is often propagated forward to some extent.\footnote{The collapse into repetition seen in many of the examples is a common effect when using argmax decoding with language models. To get qualitatively better text, other decoding methods should be used, but our goal here is to identify information being propagated through the learned recurrence mechanism.} For example, the first sample continues a description of rivers using only information passed through the recurrent module. Table~\ref{tab:pronouns} shows that the recurrent module can pass some information about what characters are in the context (mostly in the form of using the correct pronouns). The first sample, for instance, generates ``the two women'' after two women were described in the context (as well as ``the aunts'').

On the other hand, the examples also contain several discontinuities between the context and the continuation in terms of local syntax or facts, such as in the last example in Table~\ref{tab:pronouns}, where the closing quote mark which was predicted from the previous window is interpreted as an opening one. These are the types of issues that evaluation with an overlap between adjacent windows may easily address---a fact that likely accounts in part for the gap between the recurrent model with disjoint and overlapped execution shown in Table~\ref{tab:pg19}.
The examples also show new characters being invented rather than using those in the context.
The recurrent module acts as a bottleneck, and seems to have too little capacity to convey this type of information.
A higher capacity bottleneck could perhaps capture such information, but at the cost of additional compute and memory.

\section{Conclusion and Future work}
We showed that augmenting a pretrained language model with a recurrence module during fine-tuning can allow increased performance given a fixed computational budget. Our method can be similarly applied to improve the computational efficiency of pretrained models that already exist for many languages and domains, as well as for future models that will be developed. It can also allow their application to longer contexts than they were trained for, increasing their flexibility.

 There are two main extensions to this work that merit exploration. The first is increasing the capacity of the recurrence mechanism in such a way that it can pass more fine-grained information forward between windows. \citet{dai-etal-2019-transformer} have a very high capacity recurrent connection in the form of directly attending to the previous window, but this is not applicable to existing pretrained models that use absolute position embeddings. The goal would be to identify a mechanism that can pass information more effectively while still being learnable from a small fine-tuning dataset and without significantly increasing the computational cost.

The other direction is to use our recurrent method with the BERT family of models for tasks other than language modeling. Many of the tasks used to evaluate BERT-like models, such as those of the GLUE benchmark~\cite{glue}, use very short texts. On the other hand, the RACE~\cite{race} reading comprehension dataset has been used as well, and does contain some examples that exceed the 512 token context of BERT\@. Applying our method to a transformer pretrained with masked language modeling could allow application to longer contexts, or increased computational efficiency on smaller contexts.
\section*{Acknowledgments}
This material is based upon work supported by the National Science Foundation under Award Nos.~1941178 and 1941160.

\bibliography{recurrent_lm_tacl,anthology}
\bibliographystyle{acl_natbib}

\end{document}